\documentclass[runningheads]{llncs}
\usepackage{amsmath}
\usepackage{booktabs}
\usepackage{graphicx}
\usepackage{makecell}
\usepackage{amssymb}
\usepackage{multirow}
\usepackage{xcolor}
\usepackage[T1]{fontenc}
\usepackage{graphicx,verbatim}
\begin{document}
\title{MammoTracker: Mask-Guided Lesion Tracking in Temporal Mammograms}

\author{Xuan Liu\inst{1,3} \and
Yinhao Ren\inst{2,3} \and
Marc D. Ryser\inst{4} \and
Lars J. Grimm\inst{3} \and
Joseph Y. Lo \inst{1,2,3}}

\authorrunning{X. Liu et al.}

\institute{Department of Electrical and Computer Engineering, Duke University, Durham, NC \and
Department of Biomedical Engineering, Duke University, Durham, NC \and
Department of Radiology, Duke University School of Medicine, Durham,NC \and
Departments Population Health Science and Mathematics, Duke University School of Medicine, Durham,NC \\
\email{\{xuan.liu115,joseph.lo\}@duke.edu}}
    
\maketitle              

\begin{abstract}
Accurate lesion tracking in temporal mammograms is essential for monitoring breast cancer progression and facilitating early diagnosis. However, automated lesion correspondence across exams remains a challenges in computer-aided diagnosis (CAD) systems, limiting their effectiveness. We propose MammoTracker, a mask-guided lesion tracking framework that automates lesion localization across consecutively exams. Our approach follows a coarse-to-fine strategy incorporating three key modules: global search, local search, and score refinement. To support large-scale training and evaluation, we introduce a new dataset with curated prior-exam annotations for 730 mass and calcification cases from the public EMBED mammogram dataset, yielding over 20000 lesion pairs, making it the largest known resource for temporal lesion tracking in mammograms. Experimental results demonstrate that MammoTracker achieves 0.455 average overlap and 0.509 accuracy, surpassing baseline models by 8\%, highlighting its potential to enhance CAD-based lesion progression analysis. Our dataset will be available at https://gitlab.oit.duke.edu/railabs/LoGroup/mammotracker.


\end{abstract}
\section{Introduction}

Temporal analysis of mammograms across multiple consecutive exams provides valuable insights for breast cancer screening \cite{ref_6,ref_19}. An emerging area of computer-aided diagnosis (CAD) research focuses on monitoring lesion changes over time. This is achieved by establishing correspondences between current and prior mammograms, enabling the assessment of disease progression \cite{ref_18,ref_10,ref_20}. However, manually tracking lesion locations across different exams is labor-intensive. Therefore, an automated CAD framework is needed to streamline the tracking process, reducing radiologists' workload while improving efficiency and consistency. This automated lesion correspondence serves as a crucial pre-processing step for downstream temporal CAD models, enhancing both detection and classification performance.

Lesion tracking, a key component of temporal lesion monitoring, remains underdeveloped \cite{ref_23}. Existing mammogram tracking approaches primarily rely on image registration techniques to align lesion locations \cite{ref_10,ref_21,ref_11,ref_12}. However, breast tissue, being soft and deformable, undergoes slight variations across imaging instances, making rigid-body registration methods ineffective. Inspired by object tracking in natural image analysis \cite{ref_8,ref_9,ref_5,ref_13}, Siamese-based tracking models have shown success in video processing. Nevertheless, adapting such deep learning-based tracking methods for lesion tracking in global mammograms presents unique challenges. Compared to natural images, mammograms have significantly higher spatial resolution (dimensions of 2K–4K), and lesions exhibit variations in size and appearance over time. Conventional down-sampling approaches lead to substantial information loss, particularly in calcification cases, thereby negatively impacting tracking performance.

In this work, we introduce a new temporal lesion tracking dataset based on the public EMBED dataset \cite{ref_6}, providing precise lesion annotations for over 700 patients, with exams spanning up to 8 years. To address lesion tracking challenges, we propose MammoTracker, a mask-guided framework mimicking radiologists' reading behavior. It consists of (1) a global search step using registration-based approach, (2) a local search step with a mask-guided anchor-free tracking model, and (3) a score refinement step through a mask-guided similarity learning model.

\begin{figure}
\includegraphics[width=\textwidth]{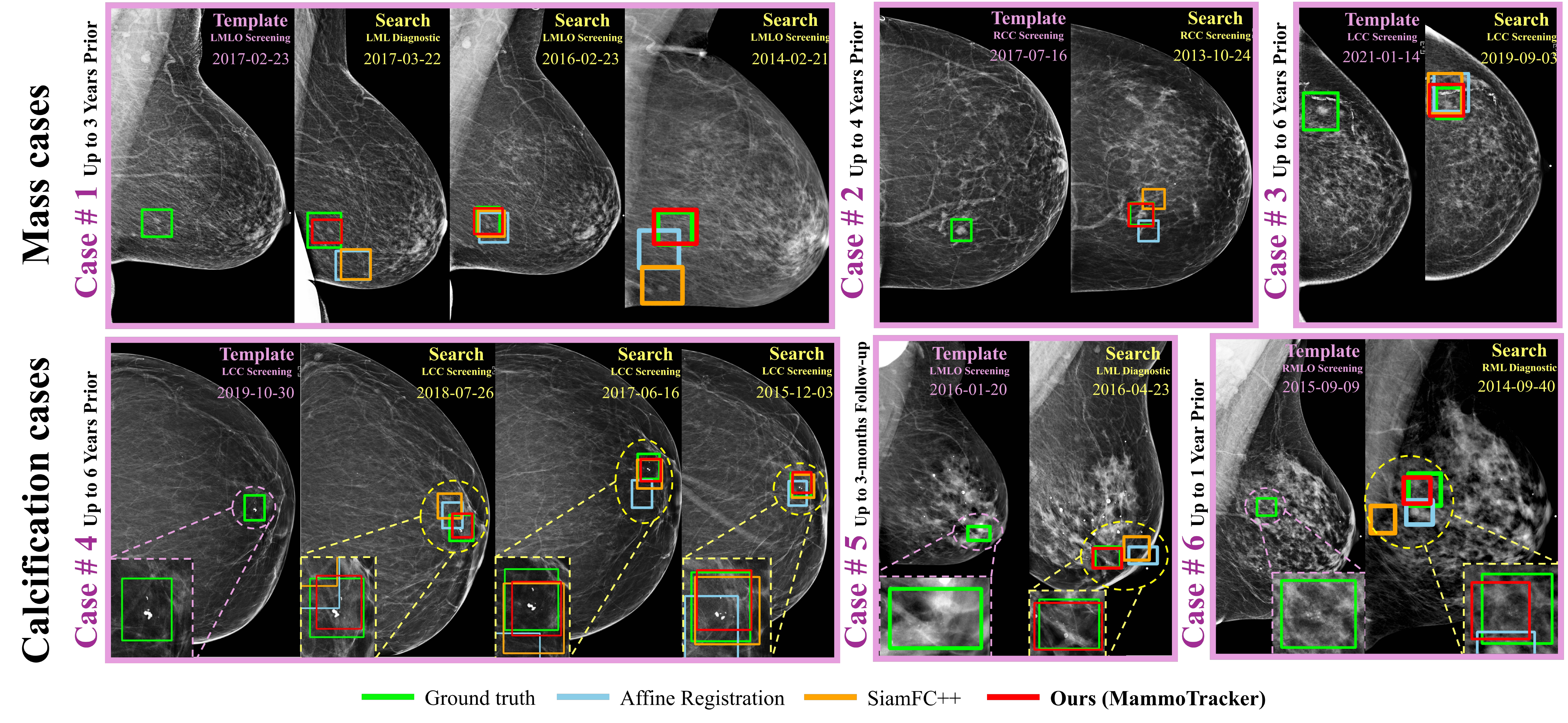}
\caption{Representative experimental results comparing MammoTracker with two baseline trackers on both screening and diagnostic images, demonstrating superior performance in scale adaptation, aspect ratio consistency, and lesion localization precision for both mass and calcification cases.} \label{fig0}
\end{figure}

We summarize our contributions as follows:
\begin{enumerate}
    \item We propose MammoTracker, a novel lesion tracking framework  to precisely identify lesion locations in temporal mammograms. As shown Fig. \ref{fig0}, MammoTracker outperforms both registration methods and deep learning-based Siamese trackers, with quantitative evaluation metrics confirming its superiority.
    \item We release the largest temporal mammogram dataset with lesion annotations for 518 mass and 212 calcification cases. With over 20000 exhaustive lesion pairs spanning up to 8 years, this dataset could be valuable for facilitating future research in temporal lesion analysis and CAD.
\end{enumerate}

\section{Related Work}
\textbf{Registration-based Approach.} Temporal lesion tracking leverages spatial consistency, as breast structure remains stable over time. This enables global registration methods (rigid, affine, Demons) for lesion alignment \cite{ref_10,ref_1}, effective for large or stable lesions but less sensitive to local deformations \cite{ref_12}. In this work, registration serves as the global search stage in our cascade tracking framework.

\noindent\textbf{Anchor-free Tracking Model.} Siamese-based tracking is widely used in visual object tracking \cite{ref_8,ref_9,ref_5,ref_13}. Inspired by anchor-free object detection \cite{ref_7}, anchor-free tracking removes predefined anchor boxes, improving efficiency and achieving state-of-the-art (SOTA) performance \cite{ref_22,ref_14}. In this work, we integrate an anchor-free tracking model as the local search component in our framework.

\noindent\textbf{Mask-guided Mechanism.} Mask-guided mechanisms, as illustrated in Fig. \ref{fig2} (b), extract robust, background-invariant features, as demonstrated in person re-identification. Chunfeng et al. \cite{ref_2} use RGB-Mask pairs to remove background noise and preserve shape information, while Honglong et al. \cite{ref_3} apply mask-guided attention for improved tracking. Inspired by this, we incorporate mask guidance into our anchor-free tracking and similarity learning models, enhancing lesion-aware feature learning.

\section{Method}

As illustrated in Fig. \ref{fig1}, the proposed MammoTracker framework consists of three main components: (1) Global Search: An affine registration-based approach aligns mammograms at the breast level, narrowing the search area for finer tracking (2) Local Search: A mask-guided anchor-free tracking model accurately localizes lesions within the refined region, improving lesion-background separation. (3) Score Refinement: A mask-guided similarity learning module refines confidence scores for predicted bounding boxes, ensuring more reliable lesion tracking.

\subsection{Global Search: Registration Alignment}
In the global search, we use affine registration \cite{ref_4} to align images by solving $ \tau_{\text{Aff}} = \arg\min \|\tau_{\text{Aff}}(I_t) - I_s\|_1 $, where $I_t$ and $I_s$ denotes the template and search images, respectively. To improve computational efficiency, both images are down sampled by a factor of 8.

\begin{figure}
\centering
\includegraphics[width=0.6\textwidth]{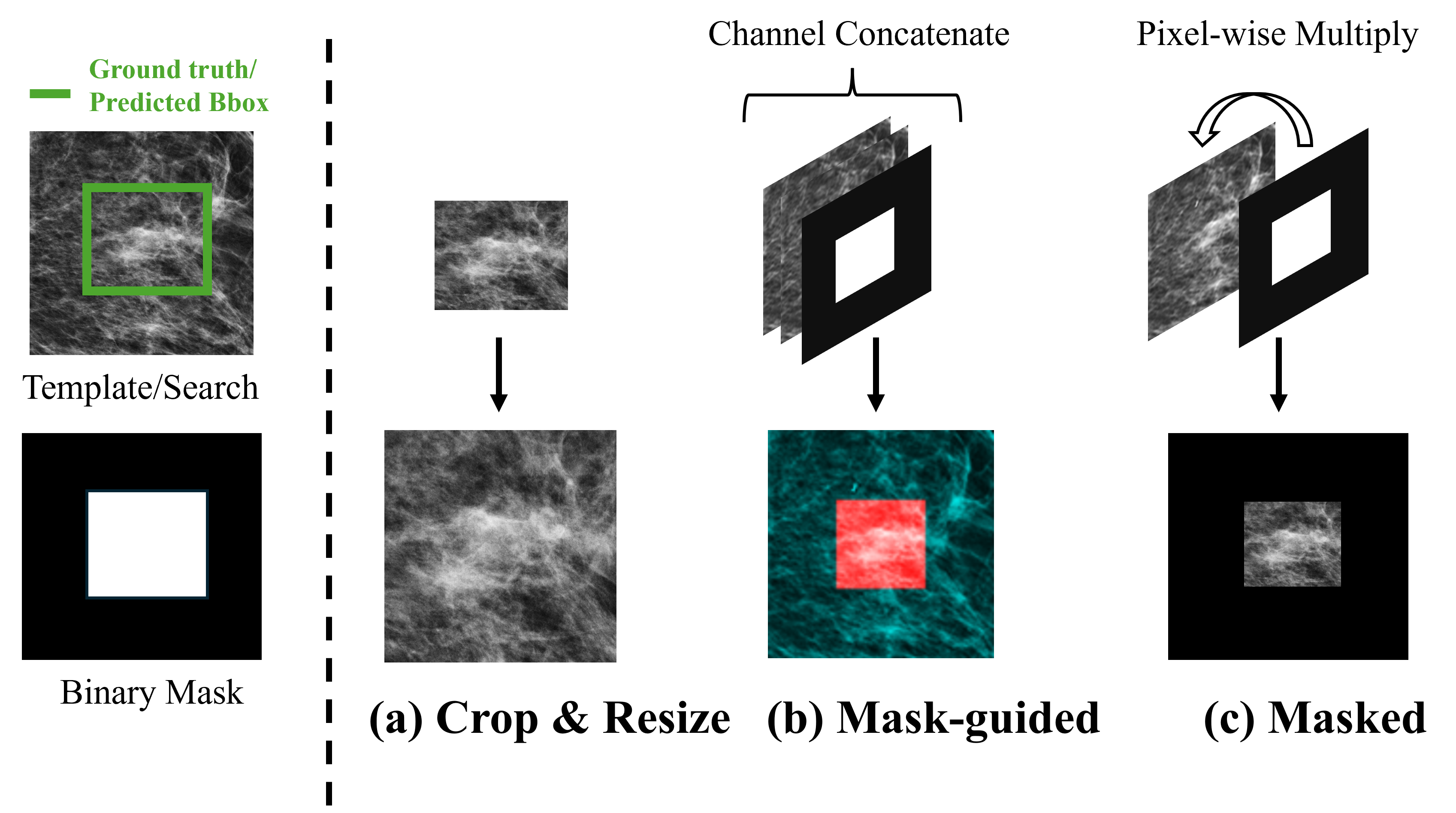}
\caption{Three different inputs. (a) Cropped \& resized input; (b) Mask-guided input; (c) Masked input.} \label{fig2}
\end{figure}

Most lesion sizes in our dataset range from 5 mm and 60 mm, with some reaching 120 mm. Based on this distribution, we define a local template size of 80 mm (\textasciitilde 1143 pixels at 0.07 mm spacing). Lesions exceeding this size retrain the registration output as the final tracking result without further refinement.

For comprehensive lesion coverage during local search, the search region size is set to 110 mm (\textasciitilde 1571 pixels at 0.07 mm spacing), capturing 97\% of lesions based on the center of the registration coordinates. Template patches are resized to 512x512 pixels, and search patches to 1024x1024 pixels, which are then used for training and inference in the cascade tracking model.

\subsection{Local Search: Mask-guided Anchor-free Tracking Model}
As shown in Fig. \ref{fig2} (b), template binary masks are generated using ground-truth bounding boxes and concatenated with corresponding template patches to form the input for tracking.

For the anchor-free tracking model, we use pre-trained MobileNetV2 \cite{ref_17} as the backbone. 
Following \cite{ref_15}, the center 7x7 template feature regions are cropped for similarity matching and a depth-wise cross-correlation layer is applied.

To suppress low-quality predicted bounding boxes, we integrate a center-ness mechanism \cite{ref_5,ref_7}. The center-ness score is computed as:
\begin{equation}
    \text{centerness} = \sqrt{\left( \frac{\min(l, r)}{\max(l, r)} \times \frac{\min(t, b)}{\max(t, b)} \right)}
\end{equation}
where l, r, t, b are distances from the predicted bounding box center to its boundaries. The final score is $cls = centerness \times classification$.

During training, focal-loss is used for classification and center-ness losses, while EIoU \cite{ref_16} is applied for regression. The total loss function is defined as:
\begin{equation}
    \text{Loss} = \lambda_1 L_{\text{classification}} + \lambda_2 L_{\text{centerness}} + \lambda_3 L_{\text{reg}}
\end{equation}

\begin{figure}
\centering
\includegraphics[width=\textwidth]{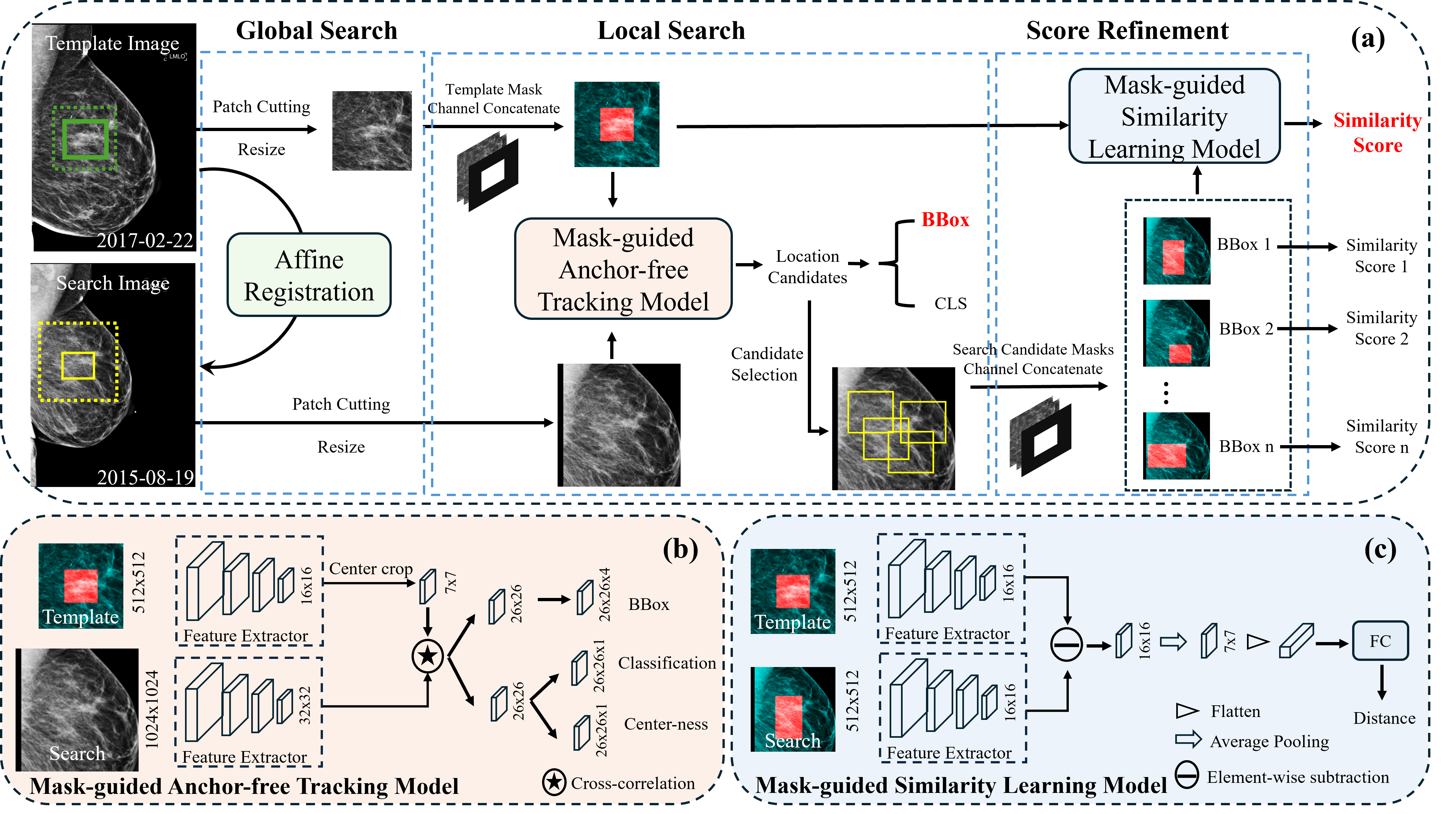}
\caption{(a) Overall framework of MammoTracker. (b) Structure of mask-guided anchor-free tracking model. (c) Structure of mask-guided similarity learning model.} \label{fig1}
\end{figure}

\noindent Where the regression loss is formulated as:
\begin{equation}
     L_{\text{reg}} = L_{\text{EIoU}} = L_{\text{IoU}} + L_{\text{dis}} + L_{\text{asp}} =  1 - \text{IoU} + \frac{\rho^2(b, b^{gt})}{(w^c)^2 + (h^c)^2} + \frac{\rho^2(w, w^{gt})}{(w^c)^2} + \frac{\rho^2(h, h^{gt})}{(h^c)^2}
\end{equation}
Where $b$ and $b^{gt}$ are bounding box centers, and $w^c$ and $h^c$ denote the smallest enclosing box dimensions.

\subsection{Score Refinement: Mask-guided Similarity Learning Model}
We observe that the local search model is limited when high-IoU bounding boxes receive low cls scores. To address this, we introduce a mask-guided similarity learning model, as illustrated in Fig. \ref{fig1}, to refine confidence scores by learning complex lesion patterns. Predicted boxes undergo non-maximum suppression (NMS) (IoU > 0.7), retaining those with cls > 0.05 for similarity learning.

The model generates binary masks from predicted boxes, concatenating them with search patches, as shown in Fig. \ref{fig2} (b). To optimize efficiency, all patches are resized to 512x512 pixels. IoU-based distance is used as 0 if IoU > 0.5, 1 if IoU < 0.3, and ignored in training but used in inference otherwise.

Since most predicted search bounding boxes correspond to negatives, we mitigate class imbalance problem by sampling negatives at twice the rate of positives. Additionally, subtraction-based feature distance computation is used, outperforming concatenation and cosine similarity. The model is trained with binary cross-entropy loss, with the final similarity score defined as $Similarity\_Score = 1 - distance$. The final output of MammoTracker combines predicted bounding boxes from the local tracking model with their corresponding similarity scores, ensuring improved lesion tracking accuracy.

\section{Experiments and Results}
\subsection{Dataset and Experiment Setup}
\textbf{Dataset.} EMBED is a large-scale publicly available mammogram dataset containing both screening and diagnostic images with a maximum follow-up period of 8 years \cite{ref_6}. However, lesion annotations are primarily available for the latest study dates, with limited prior annotations. Therefore, we manually annotate lesion locations at each prior time point for 730 cases within the 20\% open subset, using the provided region of interest (ROI) annotations as references. The curated dataset comprises approximately 70\% screening and 30\% diagnostic images, totaling 20426 lesion pairs for training and tests in this study. Table \ref{tab1} summarizes the detailed mass/calcification and train/test split. All images are rescaled to a reference pixel spacing of 0.07 mm x 0.07 mm.

\begin{table}[h]
    \centering
    \caption{Comparison of training and testing datasets with different lesion types, where exhaustive lesion pair is set as the collection of all unique lesion-to-lesion combinations derived from each case.}
    \label{tab1}
    \resizebox{0.8\textwidth}{!}{ 
    \renewcommand{\arraystretch}{1.} 
    \begin{tabular}{lccc|ccc}
        \toprule
        & \multicolumn{3}{c|}{\textbf{Train}} & \multicolumn{3}{c}{\textbf{Test}} \\
        & \textbf{Case} & \textbf{View} & \textbf{Pair} & \textbf{Case} & \textbf{View} & \textbf{Pair} \\
        \midrule
        Mass            & 352  & 625  & 8062  & 166  & 317  & 4688  \\
        Calcification   & 156  & 329  & 5690  & 56   & 120  & 1986  \\
        Train/Test Total & 518  & 954  & 13752 & 212  & 437  & 6674  \\
        \midrule
        \textbf{Total} & \textbf{Case} & \multicolumn{2}{c}{\textbf{View}} & \multicolumn{3}{c}{\textbf{Exhaustive Lesion Pair}} \\
        & 730  & \multicolumn{2}{c}{1391} & \multicolumn{3}{c}{20426} \\
        \bottomrule
    \end{tabular}
    }
\end{table}

\noindent\textbf{Evaluation metrics.} Following natural image evaluation practices \cite{ref_8}, we assess our framework using five metrics. Average overlap (AO) measures the mean IoU across all lesion pairs, while accuracy represents the mean IoU for successful tracking. Robustness evaluates the tracking failure ratio, and average center point L2 distance computes the Euclidean distance (in mm) between ground-truth and predicted bounding box centers. The success plots show the proportion of successfully tracked pairs across IoU thresholds (0 to 1), with the area under the curve (AUC) serving as a comprehensive ranking metric.

\noindent\textbf{Implementation Details.} The proposed framework is implemented using TensorFlow 2.2 and trained on four NVIDIA 2080 Ti GPUs. The Adam optimizer is used with a learning rate of 0.00005 for both the tracking and similarity learning models. The tracking model is trained for 50 epochs with a batch size of 16, while the similarity learning model is trained for 5 epochs with a batch size of 8. The affine registration method follows the settings described in \cite{ref_4}. 

\begin{table}[h]
    \centering
    \caption{MammoTracker comparisons on the test dataset.}
    \label{tab2}
    \resizebox{0.8\textwidth}{!}{ 
    \begin{tabular}{l|l|c|c|c|c}
        \toprule
        \makecell{\textbf{Lesion} \\ \textbf{Type}} & \textbf{Method} & \textbf{AO $\uparrow$} & \textbf{\scriptsize{Accuracy}$\uparrow$} & \textbf{\scriptsize{Robustness}$\downarrow$} & \makecell{\textbf{\scriptsize{L2 distance$\downarrow$}} \\ \textbf{\scriptsize (mm)}} \\
        \midrule
        \multirow{4}{*}{Mass}  
            & Affine \cite{ref_4}  & 0.389 & 0.430 & 0.095 & 12.694 \\
            & SiamFC++ \cite{ref_5} & 0.424 & 0.469 & \textbf{0.094} & 11.966 \\
            & \makecell[l]{Mask-guided Tracking}  & 0.453 & 0.501 & 0.097 & 11.535 \\
            \cmidrule(lr){2-6}
            & \textbf{MammoTracker}  & \textbf{0.467} & \textbf{0.516} & 0.095 & \textbf{11.057} \\
        \midrule
        \multirow{4}{*}{Calc}  
            & Affine \cite{ref_4}  & 0.338 & 0.404 & 0.165 & 13.761 \\
            & SiamFC++ \cite{ref_5} & 0.410 & 0.471 & \textbf{0.130} & 11.794 \\
            & \makecell[l]{Mask-guided Tracking}  & 0.412 & 0.475 & 0.133 & 11.716 \\
            \cmidrule(lr){2-6}
            & \textbf{MammoTracker}  & \textbf{0.425} & \textbf{0.490} & 0.133 & \textbf{11.239} \\
        \midrule
        \multirow{4}{*}{Total}  
            & Affine \cite{ref_4}  & 0.374 & 0.423 & 0.116 & 13.011 \\
            & SiamFC++ \cite{ref_5} & 0.420 & 0.469 & \textbf{0.105} & 11.915 \\
            & \makecell[l]{Mask-guided Tracking}  & 0.441 & 0.494 & 0.108 & 11.588 \\
            \cmidrule(lr){2-6}
            & \textbf{MammoTracker}  & \textbf{0.455} & \textbf{0.509} & 0.107 & \textbf{11.111} \\
        \bottomrule
    \end{tabular}
    }
\end{table}

\subsection{Model Comparison}
We compare our proposed MammoTracker framework against affine registration \cite{ref_4} and SiamFC++ \cite{ref_5}, representing registration-based and anchor-free tracking, respectively. Table \ref{tab2} presents quantitative results. First, our mask-guided anchor-free tracking model outperforms the SiamFC++ baseline, where template patches are cropped and resized to 512x512 pixels, as illustrated in Fig. \ref{fig2} (a). AO improves from 0.420 to 0.441 (↑5.0\%) and accuracy increases from 0.469 to 0.49 (↑4.5\%), primarily due to gains in mass lesion tracking, where AO and accuracy increase by 6.8\% and L2 distance decreases by 3.6\%.

\begin{figure}
\centering
    \includegraphics[scale=0.28]{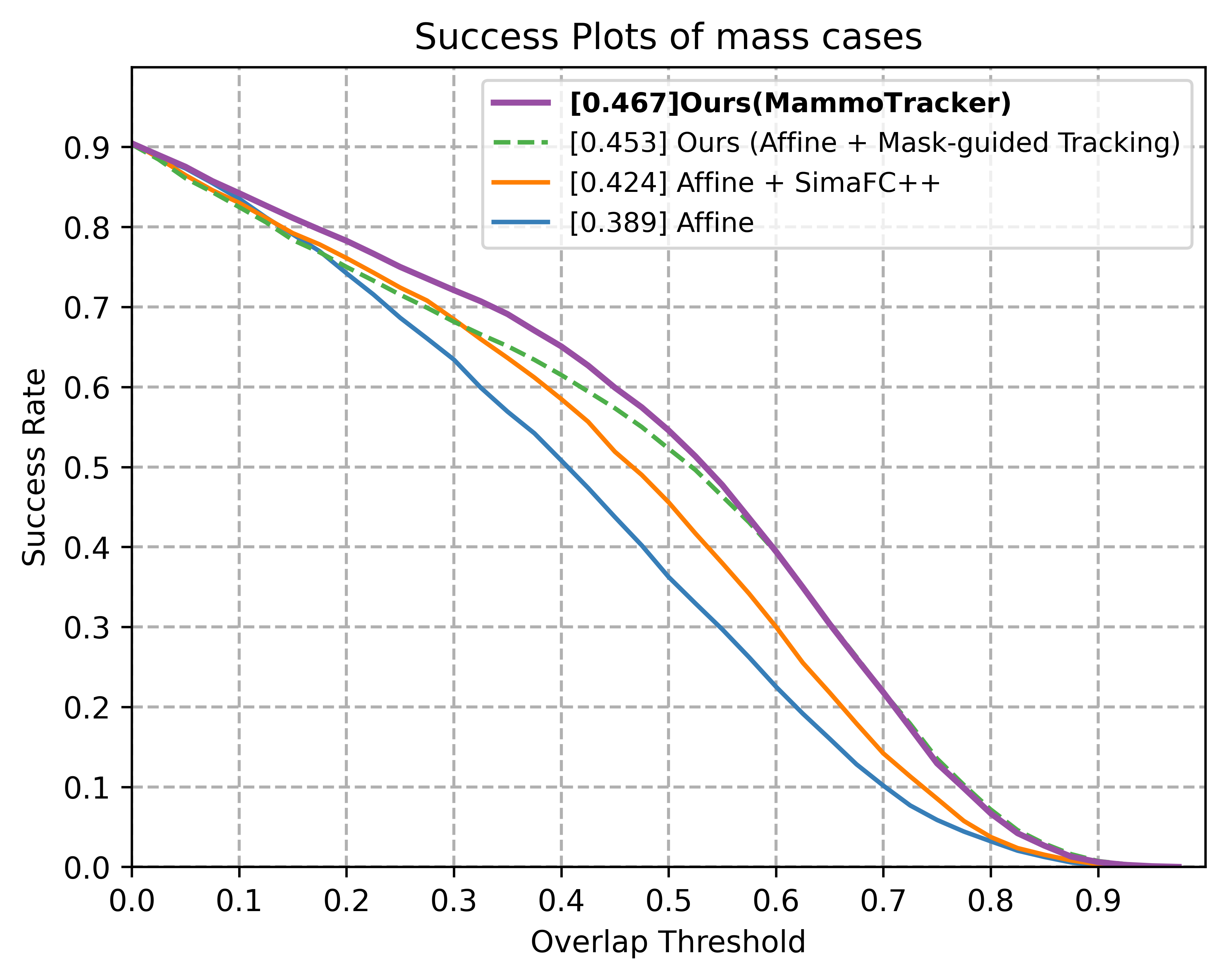} \hspace{-0.5em}
    \includegraphics[scale=0.28]{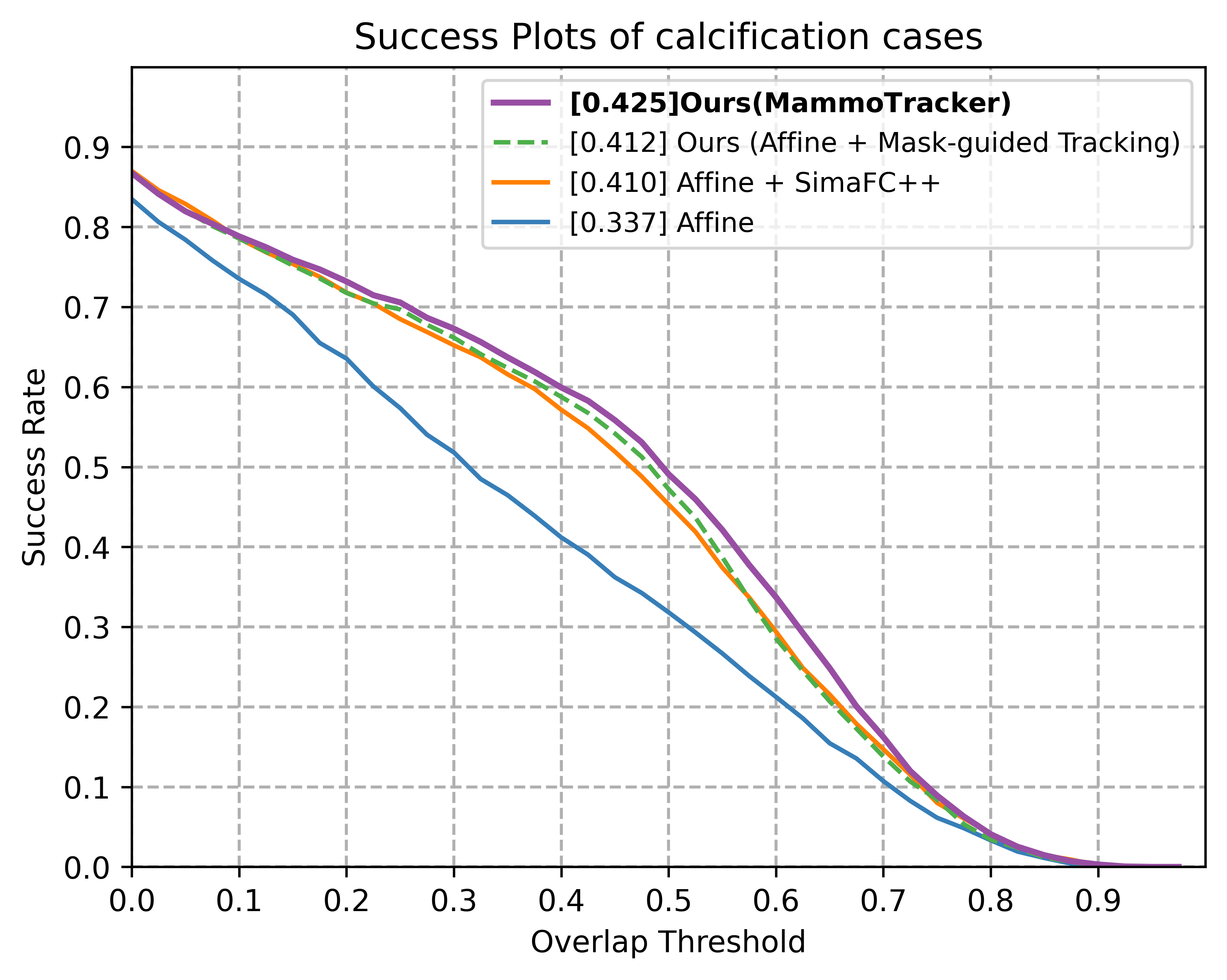} \hspace{-0.5em}
    \includegraphics[scale=0.28]{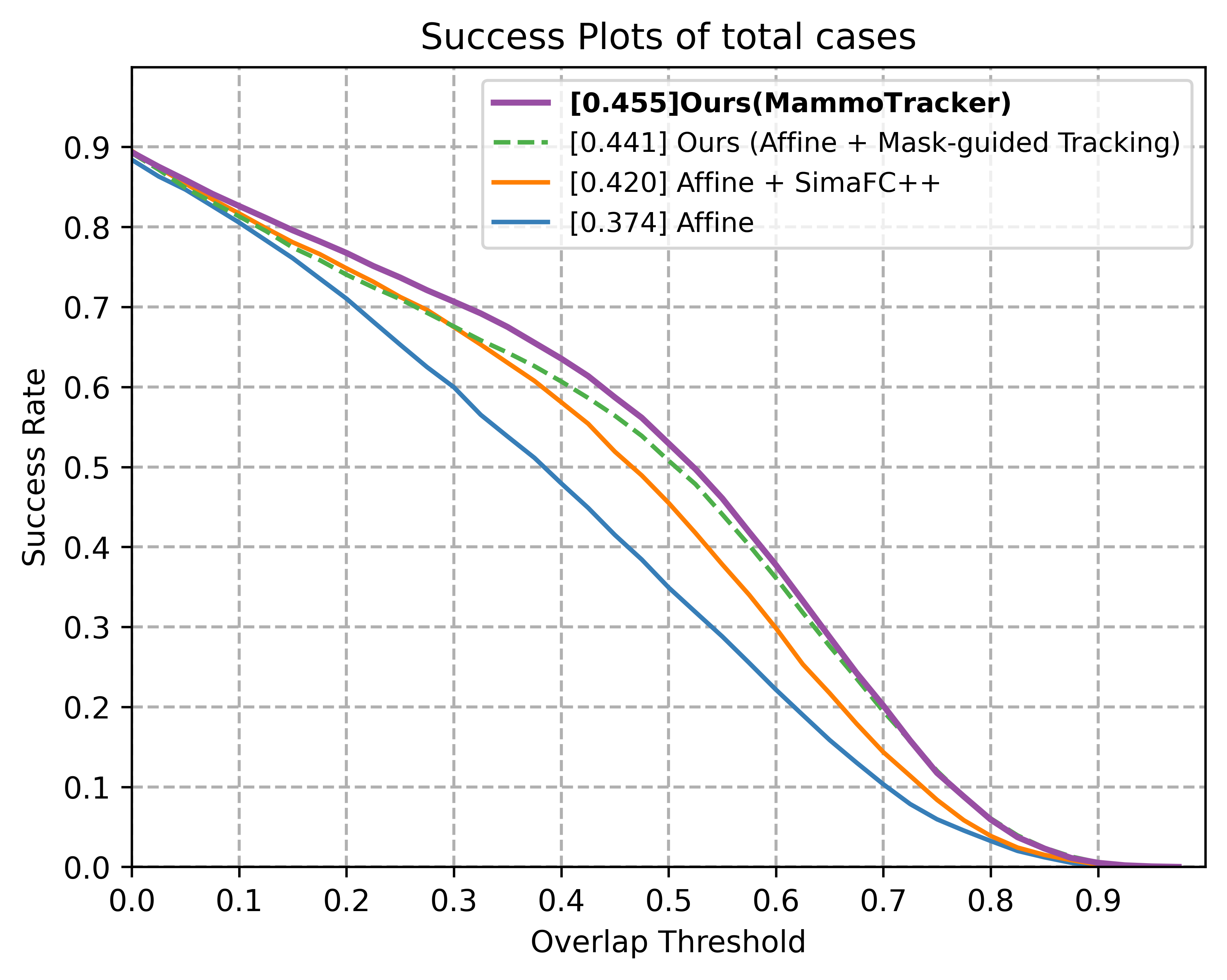}
    \caption{Success plots show a comparison of our tracker with others in different lesion types. From left to right: (a) Mass Cases; (b) Calcification Cases; (c) Both mass and calcification cases.}
    \label{fig3}
\end{figure}

Next, we evaluate the full MammoTracker framework, which includes cascade mask-guided similarity learning model. As shown in Table \ref{tab2}, it achieves notable improvements across AO and accuracy. Specifically, for mass lesions, AO increases from 0.424 to 0.467 (↑10.1\%), and accuracy from 0.469 to 0.516 (↑10.0\%). For calcifications, AO rises from 0.410 to 0.425 (↑3.7\%), while accuracy improves from 0.471 to 0.490 (↑4.0\%). Additionally, L2 distance is reduced by over 5\% for both lesion types. Fig. \ref{fig3} further illustrates that MammoTracker consistently achieves higher success rates across all IoU thresholds compared to SiamFC++ for both mass and calcification lesions.

However, SiamFC++ shows slightly better robustness, indicating potential areas for improvement in failure recovery for challenging cases.

\subsection{Ablation Study}
We conducted ablation studies to evaluate the effectiveness of our proposed methods in 2 key aspects  within the mask-guided anchor-free tracking model.

\noindent\textbf{Center-ness.} We evaluate the impact of incorporating center-ness in our anchor-free tracking model. As shown in Table \ref{tab3}, we compared two approaches: (1) using only the classification score and (2) using the product of the classification and center-ness score. Results show that center-ness significantly improves AO, accuracy and L2 distance by suppressing low-quality bounding boxes, leading to more reliable tracking performance.

\noindent\textbf{Mask-guided vs Masked Template Input.} We further examine the effect of different template input types by training the tracking model using three variations: (1) crop \& resize; (2) mask-guided and (3) masked template, as shown in Fig. \ref{fig2}. As summarized in Table \ref{tab3}, the mask-guided template consistently outperforms the other methods across AO, accuracy and L2 distance metrics. This superiority can be attributed to three factors, aligning with findings from \cite{ref_2}: (1) rectangular masks effectively separate lesions from the background, enhancing discriminative feature learning; (2) the mask preserves stable bounding box shape information over time, preventing abrupt aspect ratio shifts; and (3) unlike fully masked templates, mask-guided templates retain weak background features, providing contextual cues for improved lesion localization.

\begin{table}[h]
    \centering
    \caption{Ablation study on each module in anchor-free tracking model for both mass and calcification cases.}
    \label{tab3}
    \resizebox{0.95\textwidth}{!}{ 
    \renewcommand{\arraystretch}{1.3} 
    \setlength{\tabcolsep}{4pt} 
    \begin{tabular}{c|ccc|c|c|c|c}
        \toprule
        \textbf{Centerness} & \multicolumn{3}{c|}{\textbf{Template Inputs}} & \textbf{AO $\uparrow$} & \textbf{Accuracy $\uparrow$} & \textbf{Robustness $\downarrow$} & \textbf{L2 distance $\downarrow$} \\
        & Crop \& Resize  & Mask-guided  & Masked  & & & & \textbf{(mm)} \\
        \midrule
         & \checkmark &  &  & 0.413  & 0.461  & \textbf{0.104}  & 12.024 \\
        &  & \checkmark &  & 0.431  & 0.484  & 0.110  & 11.715 \\
        &  & & \checkmark & 0.407  & 0.464  & 0.123  & 11.776 \\
        \checkmark & \checkmark &  &  & 0.420  & 0.469  & 0.105  & 11.915 \\
        \checkmark &  &  & \checkmark & 0.416  & 0.479  & 0.132  & 11.778 \\
        \midrule
        \textbf{\checkmark} &  & \textbf{\checkmark} &  & \textbf{0.441} & \textbf{0.494} & 0.108 & \textbf{11.588} \\
        \bottomrule
    \end{tabular}
    }
\end{table}

\section{Conclusion}
In this study, we introduce MammoTracker, a mask-guided lesion tracking framework for temporal mammograms that enables accurate lesion localization across multiple time points. It follows a coarse-to-fine strategy that replicates radiologists' approach to reading sequential images and identifying corresponding lesions. We also release a large-scale dataset with over 20000 tracking pairs, based on the EMBED dataset.  MammoTracker outperforms baseline models in tracking accuracy, average overlap and L2 distance. In future work, we will extend our framework to downstream CAD tasks, such as lesion detection and classification, to further enhance breast cancer diagnosis.

%
%
%
%

\end{document}